\documentclass[conference]{IEEEtran}
\IEEEoverridecommandlockouts

\usepackage{cite}
\usepackage{amsmath,amssymb,amsfonts}
\usepackage{algorithmic}
\usepackage{graphicx}
\usepackage{textcomp}
\usepackage{xcolor}
\usepackage[hyphens]{url}
\usepackage{fancyhdr}
\usepackage{hyperref}
\usepackage[normalem]{ulem}
\usepackage{textgreek}
\usepackage{multirow}
\usepackage{makecell}
\usepackage{xspace}
\usepackage[caption=false]{subfig}
\usepackage{colortbl}
\usepackage{tabulary}
\usepackage{etoolbox}
\usepackage{booktabs}
\usepackage{mathtools}
\usepackage{pifont}
\usepackage{color}
\usepackage[dvipsnames]{xcolor}
\usepackage{tikz}
\usepackage{tcolorbox}
\usepackage{ifthen}
\usepackage{comment}
\usepackage{enumitem}
\usepackage{soul}

\newcommand{\circlednum}[1]{\ignorespacesafterend
  \tikz[baseline=-0.8ex] \node[draw, circle, fill=black, text=white, inner sep=0.05mm]{#1};\ignorespacesafterend
}
\robustify{\circlednum}

\newcommand{\wcirclednum}[1]{\ignorespacesafterend
  \tikz[baseline=-0.8ex] \node[draw, circle, fill=white, text=black, inner sep=0.05mm]{#1};\ignorespacesafterend
}
\robustify{\wcirclednum}
\newcommand{\fig}[1]{Figure~\ref{#1}}
\newcommand{\sect}[1]{Section~\ref{#1}}
\newcommand{\tab}[1]{Table~\ref{#1}}

\def\BibTeX{{\rm B\kern-.05em{\sc i\kern-.025em b}\kern-.08em
    T\kern-.1667em\lower.7ex\hbox{E}\kern-.125emX}}

\begin{document}

\title{\huge The Cost of Dynamic Reasoning: Demystifying AI Agents and \\Test-Time Scaling from an AI Infrastructure Perspective
}

\author{\IEEEauthorblockN{Jiin Kim}
\IEEEauthorblockA{
KAIST\\
jiin.kim@kaist.ac.kr}
\and
\IEEEauthorblockN{Byeongjun Shin}
\IEEEauthorblockA{
KAIST\\
byeongjun.shin@kaist.ac.kr}
\and
\IEEEauthorblockN{Jinha Chung}
\IEEEauthorblockA{
KAIST\\
jinha.chung@kaist.ac.kr}
\and
\IEEEauthorblockN{Minsoo Rhu}
\IEEEauthorblockA{
KAIST\\
mrhu@kaist.ac.kr}
}

\maketitle

\begin{abstract}
Large-language-model (LLM)–based AI agents have recently showcased impressive versatility by employing dynamic reasoning, an adaptive, multi-step process that coordinates with external tools. This shift from static, single-turn inference to agentic, multi-turn workflows broadens task generalization and behavioral flexibility, but it also introduces serious concerns about system-level cost, efficiency, and sustainability. This paper presents the first comprehensive system-level analysis of AI agents, quantifying their resource usage, latency behavior, energy consumption, and datacenter-wide power consumption demands across diverse agent designs and test-time scaling strategies. We further characterize how AI agent design choices, such as few-shot prompting, reflection depth, and parallel reasoning, impact accuracy-cost tradeoffs. Our findings reveal that while agents improve accuracy with increased compute, they suffer from rapidly diminishing returns, widening latency variance, and unsustainable infrastructure costs. Through detailed evaluation of representative agents, we highlight the profound computational demands introduced by AI agent workflows, uncovering a looming sustainability crisis. These results call for a paradigm shift in agent design toward compute-efficient reasoning, balancing performance with deployability under real-world constraints.
\end{abstract}


\section{Introduction}

Recent progress in large language models (LLMs) has shifted from scaling model size or pretraining data to improving inference-time behavior, a direction known as \textit{test-time scaling}\cite{snell2024scalingllmtesttimecompute,muennighoff2025s1simpletesttimescaling}. Test-time scaling is designed to enhance model performance by allocating additional computation during inference without modifying the model's parameters. This includes techniques such as Chain-of-Thought~\cite{cot}, Tree-of-Thought~\cite{tot}, and others~\cite{touvron2023llama2openfoundation,self_cot,least_cot,press2023measuring,graph_of_thought}. These approaches promote more deliberate and interpretable \emph{reasoning} within the LLM, enabling it not only to recognize patterns but also to derive conclusions, generate explanations, and solve tasks that require step-by-step logic.

\begin{figure}[t]
    \centering
    \includegraphics[width=0.485\textwidth]{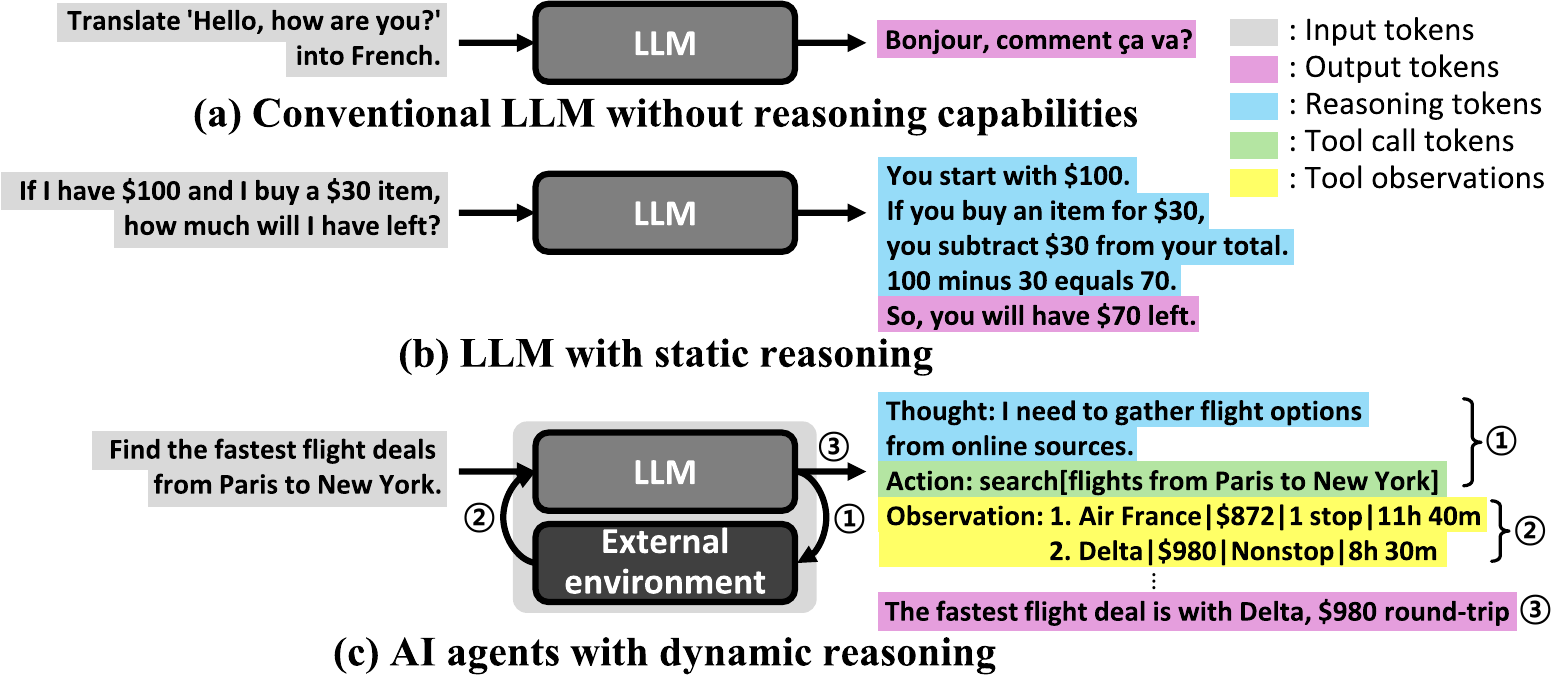}
    % \vspace{-1.8em}
    \caption{
    Overview of reasoning strategies in LLM-based systems.
    (a) Conventional LLMs map inputs directly to outputs in a single forward pass, with no explicit intermediate reasoning.
    (b) Reasoning‑enhanced LLMs internally create intermediate steps (e.g., sampling alternative responses or extending token sequences) to deepen or diversify their thought process.
    (c) AI agents augment this reasoning by
    \wcirclednum{1} planning and invoking external tools, \wcirclednum{2} observing the outcomes and adapting their internal reasoning, and iteratively refining their decision-making until they \wcirclednum{3} generate the final answer.
    }
    \label{fig:intro_overview}
    \vspace{-0.6em}    
\end{figure}

Deploying these reasoning-enhanced LLMs, however, comes at an immense computational cost. Even in current \textbf{static reasoning} models which follow fixed input-output mappings without external tool interaction (\fig{fig:intro_overview}(a,b)), LLMs run on thousands of GPUs, whose power, cooling, and capital costs drive monthly expenses into the tens of millions of dollars~\cite{openai_cost_2023}. A single ChatGPT query is estimated to consume about ten times the electricity of a typical web search~\cite{AI_energy_footprint} and requires a substantial amount of cooling water~\cite{AI_water_footprint}. As a result, hyperscalers are investing at an unprecedented scale. Meta has already committed over \$10 billion to AI infrastructure~\cite{meta_10_billion}, and Microsoft operates custom AI supercomputers that draw tens to hundreds of megawatts. As an example, xAI’s Colossus AI supercomputer alone employs approximately 100,000 Nvidia H100 GPUs, consuming 150 megawatts in total, with the GPUs alone accounting for around 70 megawatts~\cite{xAI_collosus}. For perspective, traditional hyperscale data centers typically draw between 10 and 100 megawatts~\cite{hyperscale_datacenter}, while advanced semiconductor fabs, such as those operated by Samsung and TSMC, consume several hundred megawatts each~\cite{tsmc_fab_power,fab_power}. Analysts forecast that total AI infrastructure spending will surpass \$1 trillion within this decade~\cite{capex_forecast}, raising serious concerns about power grid sustainability and the economic viability of large-scale LLM deployment.

Given this landscape, the emergence of \emph{AI agents} powered by LLMs with \textbf{dynamic reasoning} threatens to exacerbate these already formidable infrastructure pressures dramatically. Unlike static reasoning models, dynamic reasoning represents an advanced form of test-time scaling that significantly boosts capabilities through active interaction with external environments (\fig{fig:intro_overview}(c)). Specifically, AI agents continuously plan, invoke external tools, observe outcomes, and iteratively refine their reasoning, often performing dozens of inference calls to satisfy a single user request~\cite{react,reflexion,lats}. Without substantial system-level innovations, per-request computational costs could increase by orders of magnitude, making large-scale deployment of agents economically and environmentally prohibitive. Industry leaders are already responding to these challenges: OpenAI’s planned Stargate project~\cite{openai_stargate} and Meta's next generation AI data center Hyperion~\cite{meta_hyperion_prometheus} are each projected to require multiple ``gigawatts'' of power capacity, with costs reaching hundreds of billions of dollars. Yet, despite these developments, the computer architecture community has largely focused on static LLMs, leaving the infrastructure implications of dynamic reasoning workloads underexplored.

To address this critical gap, this paper presents a rigorous, quantitative evaluation of the computational and infrastructural costs of dynamic reasoning. We systematically characterize resource utilization, latency implications, and energy demands inherent in the iterative execution patterns of AI agents.
While the serving characteristics of static reasoning LLMs are well understood within the research community, our work presents the first comprehensive, system-level characterization of agent serving costs across diverse configurations and workloads. Each component of our analysis—including the characterization of agent workflows themselves, serving performance, and the impact of test-time scaling—quantifies a distinct dimension of this cost structure, collectively building a unified understanding of its infrastructure-level implications.
Our analysis highlights the critical system-level challenges faced when deploying AI agents and identifies opportunities for optimization through architectural improvements, enhanced inference algorithms, and intelligent resource allocation strategies. To the best of our knowledge, this work is the first to provide a system-level characterization of dynamic reasoning in AI agents, grounded in quantitative analysis of end-to-end infrastructure behavior across diverse agentic workflows\footnote{Open-sourced at \href{https://github.com/VIA-Research/AgentBench}{https://github.com/VIA-Research/AgentBench.}}. A key contribution and objective of our study is to quantitatively assess the AI infrastructure cost of dynamic reasoning deployments, and to inform and caution the research community about the urgent need for sustainable, efficient design principles to bridge the gap between advanced algorithmic capabilities and practical, scalable, and sustainable deployment.

\section{Background and Motivation}
\label{sect:background}

\subsection{Definition of AI Agents}
\label{sect:llm_agent_definition}

AI agents are inference-time frameworks that extend the capabilities of LLMs by enabling multi-step reasoning, adaptive decision-making, and interaction with the external environment. Unlike conventional LLM applications that produce a single output from a static prompt, AI agents operate through iterative internal reasoning and external actions at inference time. At each iteration, the agent may generate an intermediate reasoning result, call an external tool (e.g., search engine, calculator, or code interpreter), and incorporate the output into its subsequent decisions. This process allows the agent to retrieve missing information and refine its strategy \emph{dynamically} in response to evolving task demands. While this adaptivity enhances the ability to handle complex and open-ended problems, it also leads to variability in the LLM calls, tool usage patterns, and overall computational cost.

\subsection{Core Components and Workflows of AI Agents}
\label{sect:llm_agent_components}

\begin{figure}[t!] \centering
\includegraphics[width=0.48\textwidth]{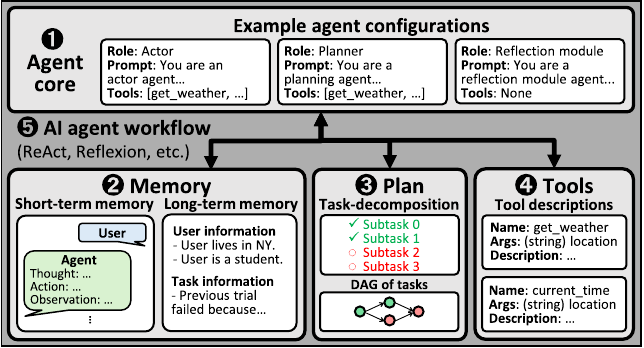}\\
\vspace{-0.4em}    
\caption{\small Overview of AI agent structure.
}
\label{fig:agent_overview}
\vspace{-0.6em}    
\end{figure}

As illustrated in \fig{fig:agent_overview}, AI agents generally consist of four core components (\emph{agent core, memory, plan}, and \emph{tools}), and \emph{AI agent workflows} interconnect these core components through iterative interactions. These workflows orchestrate how the components dynamically collaborate, enabling the agent’s adaptive behaviors.

The \circlednum{1}~\emph{agent core} is the central component responsible for \emph{advanced reasoning}, powered by one or more LLMs configured in specific ``roles''. These roles typically include an \emph{actor}, which determines the agent’s next action; a \emph{planner}, which decomposes high-level goals into subtasks; and a \emph{reflection module}, which evaluates prior reasoning steps and tool interaction trajectories to guide future decisions.

This core reasoning capability of the AI agent is further supported by \emph{memory}, \emph{plan}, and \emph{tools}. \circlednum{2} \emph{Memory} plays a critical role in enabling the agent to maintain continuity across reasoning steps by storing short-term interaction traces as well as long-term knowledge, including user preferences or experience from past interactions. 
\circlednum{3} \emph{Plan} organizes the agent’s objective into a sequence of subtasks or a directed acyclic graph (DAG) of interdependent actions. By maintaining an explicit plan, the agent can prioritize actions, track progress, and make forward-looking decisions that align with the overall task structure.
\circlednum{4} \emph{Tools} extend the agent’s capabilities beyond text generation by enabling interaction with external environments. At each step, the agent analyzes its current context, generates a structured command specifying the desired tool and input, executes the tool call, and incorporates the resulting output into its context. This output is then used to guide the next stage of reasoning.

Finally, \circlednum{5}~\emph{AI agent workflow} defines how an agent leverages interactions among the four core components iteratively to carry out reasoning and coordinate actions. AI agents implement their own distinct workflows, reflecting different coordination patterns among components. These workflows can be broadly decomposed into two phases: (1) \emph{LLM inference phase}, where the agent performs internal reasoning tasks such as action generation, planning, or reflection; and (2) \emph{tool use phase}, where the agent interacts with external environments using tools. These two phases alternate iteratively, forming the backbone of AI agentic systems.

\subsection{Test-Time Scaling in AI Agents}

Test-time scaling refers to methods that improve the reasoning performance of pretrained LLMs at inference time by increasing the amount of computation used for inference without modifying model parameters~\cite{cot,tot,self_cot,least_cot,snell2024scalingllmtesttimecompute}. Representative techniques include Chain-of-Thought~\cite{cot}, which guides the model to produce intermediate reasoning steps through carefully crafted prompts, and Tree-of-Thought~\cite{tot}, which expands the reasoning space by exploring multiple reasoning paths. These approaches guide the model to perform step-by-step reasoning, effectively leveraging its internal reasoning capabilities while keeping its parameters fixed.

AI agents build upon this paradigm by implementing test-time reasoning not through prompt design alone, but through multi-step decision-making that integrates tool use and maintenance of intermediate reasoning state. Unlike conventional prompt-based methods that operate within a \emph{static} input-output mapping, agents \emph{dynamically} coordinate multiple model invocations and tool interactions, adapting their behavior based on intermediate outcomes. This form of \emph{dynamic reasoning} enables agents to respond to new information, revise prior decisions, and handle real-time tasks involving external environments.
As such, AI agents redefine test-time scaling by moving beyond conventional inference approaches that rely solely on the model’s internal reasoning abilities.
This paradigm shift introduces new challenges in efficiency, latency, and resource management, highlighting the need for a system-level analysis of the behavior of AI agents.

\subsection{Motivation}

Unlike conventional single-turn LLM inference where computation is bounded to a single forward pass, agentic execution involves dynamically evolving control flows, multiple rounds of LLM inference, and external tool interactions. These behaviors introduce profound challenges at the systems and infrastructure level, incurring significant compute overhead, amplifying memory pressure, and introducing unpredictable latency and resource utilization patterns.

Despite these operational complexities, prior research on AI agents has largely focused on improving task success rates and qualitative reasoning behavior~\cite{react, reflexion, lats}, with little attention paid to its deployment costs. Questions central to the deployment and scaling of such agents remain largely unexamined. Consequently, existing architecture and systems optimizations for LLMs, which target static, single-pass workloads, may fall short in capturing or addressing the dynamic and iterative characteristics unique to AI agents.

This paper is motivated by the urgent need to fill this gap. To the best of our knowledge, this work is the first to present a rigorous, system-level characterization of AI agents, grounded in quantitative measurement across diverse agent designs and tasks. We argue that without a principled understanding of the system-level implications of dynamic reasoning, the community risks building infrastructure optimized for yesterday’s workloads. A systems-oriented perspective is therefore critical to guide the design of sustainable, efficient, and scalable serving infrastructures. Our study takes this first step by analyzing the computational and infrastructural costs of deploying AI agents in practice, providing actionable insights for future architecture and systems co-design.

\section{Methodology}

Our analysis considers a representative set of AI agents and benchmarking workloads, covering diverse agent workflows and agentic task characteristics.

\begin{scriptsize}
\begin{table}[t]
\centering
\caption{Comparison of AI agents.}
\scriptsize
\begin{tabular}{cccccc}
\hline
\textbf{Agent} & \textbf{Reasoning} & \makecell[c]{\textbf{Tool}\\ \textbf{Use}} & \textbf{Reflection} & \makecell[c]{\textbf{Tree}\\ \textbf{Search}} & \makecell[c]{\textbf{Structured}\\ \textbf{Planning}} \\ \hline\hline
\textbf{CoT}~\cite{cot} & O & X & X & X & X \\ 
\textbf{ReAct}~\cite{react} & O & O & X & X & X \\ 
\textbf{Reflexion}~\cite{reflexion} & O & O & O & X & X \\ 
\textbf{LATS}~\cite{lats} & O & O & O & O & X \\ 
\textbf{LLMCompiler}~\cite{llmcompiler} & O & O & O & X & O \\ \hline
\end{tabular}
\label{tab:agents}
\end{table}
\end{scriptsize}

\begin{figure}[t!] \centering
\captionsetup[subfloat]{captionskip=0.1em}
\subfloat[CoT]{\includegraphics[width=0.19\textwidth]{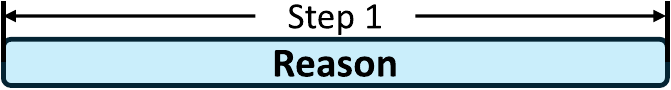}}\\
\subfloat[ReAct]{\includegraphics[width=0.24\textwidth]{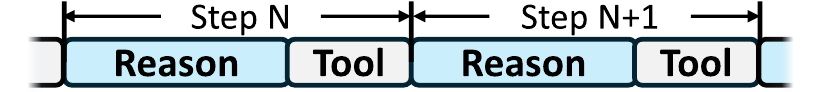}}\\
\subfloat[Reflexion]{\includegraphics[width=0.32\textwidth]{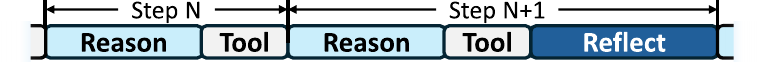}}\\
\subfloat[LATS]{\includegraphics[width=0.48\textwidth]{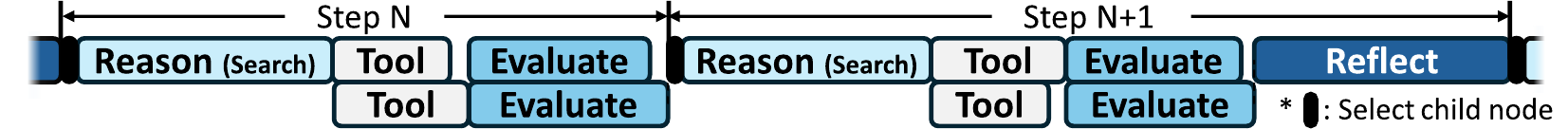}}\\
\subfloat[LLMCompiler]{\includegraphics[width=0.39\textwidth]{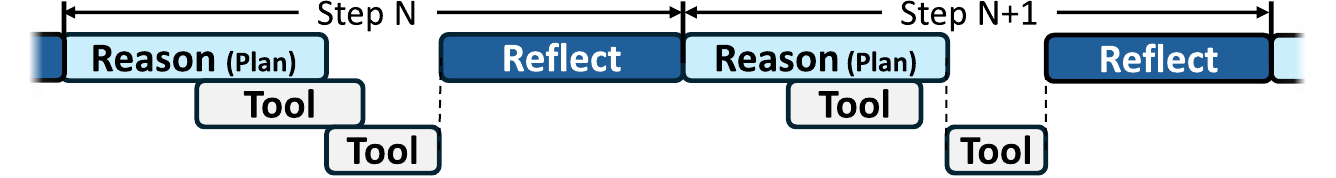}}
\label{fig:agent_timeline_llmcompiler}
\\
\caption{Execution timeline of each AI agent.}
\label{fig:agent_timeline}
\vspace{-0.8em}
\end{figure}

\textbf{AI agent workflows.} 
We investigate five representative agents: Chain-of-Thought (CoT)~\cite{cot}, ReAct~\cite{react}, Reflexion~\cite{reflexion}, Language Agent Tree Search (LATS)~\cite{lats}, and LLMCompiler~\cite{llmcompiler}. These agents were selected to cover a wide range of reasoning strategies, tool integrations, and planning mechanisms. Table \ref{tab:agents} summarizes the presence or absence of five key capabilities across each agent.

\begin{itemize}[leftmargin=1em]
    \item \textbf{Reasoning.} All agents considered in this study employ a reasoning mechanism. Among them, CoT operates purely through internal reasoning without the use of any external tool (\fig{fig:agent_timeline}(a)). As a baseline for comparison, CoT-style static reasoning approaches are considered within the broader definition of AI agents, despite their lack of external interactions with tools.
    
    \item \textbf{Tool use.} Tool use differentiates purely language-based agents from those capable of interacting with the external environment. This functionality enables agents to access real-time data or perform non-linguistic operations.

    \item \textbf{Reflection.} Reflection allows agents to evaluate past decisions and revise strategies accordingly. Reflective agents effectively manage \emph{long-term memory} by abstracting past trajectories into reflections. While ReAct agents simply repeat reasoning and tool usage (\fig{fig:agent_timeline}(b)), Reflexion, the most fundamental reflective agent, enhances adaptability by periodically incorporating self-evaluation and refinement through reflection (\fig{fig:agent_timeline}(c)).

    \item \textbf{Tree search.} LATS (\fig{fig:agent_timeline}(d)) leverages Monte Carlo Tree Search~\cite{mcts} to simulate multiple branches of reasoning and action, allowing the agent to evaluate different candidate paths before making a decision. By simulating multiple possible future paths, the agent can make more informed decisions and select optimal action sequences.

    \item \textbf{Structured planning.} LLMCompiler incorporates a structured multi-step planning and streaming for asynchronous task execution to minimize latency. During the planning phase, LLMCompiler analyzes task dependencies and constructs a DAG that organizes future tool calls into an execution plan. This enables multiple dependent tool calls to be generated within a single LLM invocation. As the plan is constructed, intermediate tool calls are streamed to the execution stage, allowing the scheduler to overlap planning and tool calls via asynchronous execution. Together, these features can help reduce repeated reasoning and lower end-to-end latency (\fig{fig:agent_timeline}(e)).
\end{itemize}

In general, we utilized the official open-source implementations provided by the original authors of these agent workflows~\cite{react_github,reflexion_github,lats_github,llmcompiler_github}. Each AI agent is adapted to support our evaluation framework and benchmarks. For LATS, we further optimized its implementation to support concurrent LLM inference and parallel tool invocation because the original version~\cite{lats_github} executes these operations sequentially, aggravating end-to-end latency.

\begin{scriptsize}
\begin{table}[t]
\centering
\caption{Description of benchmarks.}
\scriptsize
\begin{tabular}{ccc}
\hline
\textbf{Benchmark} & \textbf{Property} & \textbf{Description} \\ \hline\hline
\multirow{3}{*}{\textbf{HotpotQA}~\cite{hotpotqa}} 
    & \textbf{Task} & Multi-hop question answering \\ \cline{2-3}
    & \textbf{Tool} & Wikipedia APIs (search, lookup keywords)\\ \cline{2-3}
    & \textbf{Agent} & CoT, ReAct, Reflexion, LATS, LLMCompiler \\ \hline
\multirow{3}{*}{\textbf{WebShop}~\cite{webshop}} 
    & \textbf{Task} & Online shopping \\ \cline{2-3}
    & \textbf{Tool} & Interactive web navigation (search, click) \\ \cline{2-3}
    & \textbf{Agent} & ReAct, Reflexion, LATS, LLMCompiler \\ \hline
\multirow{3}{*}{\textbf{MATH}~\cite{math}} 
    & \textbf{Task} & Math problem solving \\ \cline{2-3}
    & \textbf{Tool} & Wolfram Alpha API, Python-based calculator \\ \cline{2-3}
    & \textbf{Agent} & CoT, ReAct, Reflexion, LATS \\ \hline
\multirow{3}{*}{\textbf{HumanEval}~\cite{humaneval}} 
    & \textbf{Task} & Programming \\ \cline{2-3}
    & \textbf{Tool} & Executing self-generated test code \\ \cline{2-3}
    & \textbf{Agent} & CoT, ReAct, Reflexion, LATS \\ \hline
\end{tabular}
\label{tab:benchmarks}
\end{table}
\end{scriptsize}

\textbf{Benchmarks.}
We select four popular benchmarks representative of various downstream agentic tasks, whose descriptions are summarized in \tab{tab:benchmarks}. 
HotpotQA~\cite{hotpotqa} is a question-answering benchmark that assesses the agent's ability to accurately retrieve relevant evidence to answer multi-hop knowledge-intensive questions. We provide the Wikipedia APIs~\cite{wikipedia_api} as tools to solve these questions.
WebShop~\cite{webshop} is a web-shopping benchmark where agents find the best-fit item that meets the given conditions. The agent is given web navigation tools to browse WebShop.
MATH~\cite{math} is a benchmark suite of mathematics problems across various domains. Agents are equipped with access to the Wolfram Alpha API~\cite{wolfram_alpha_api} for solving complex equations, as well as a Python-based calculator for simple numerical computations.
HumanEval~\cite{humaneval} evaluates the programming capability of agents. In our setup, agents are equipped with a Python execution tool that allows them to validate the generated solutions by executing self-written test code.
In addition to these agentic benchmarks, we utilize a \emph{non-agentic} dataset, which is the ShareGPT dataset~\cite{sharegpt}, to model conventional chatbot-like LLMs, characterized by single-turn LLM inference without iterative interactions with the external environments. ShareGPT contains a collection of real conversations between users and ChatGPT~\cite{chatgpt}, capturing standard interactive dialogue scenarios.

It is worth pointing out that some ``AI agent vs. benchmark'' pairs are omitted if the agent is not suitable for solving the target task. For example, CoT is excluded from WebShop since it cannot interact with the shopping webpage. Similarly, LLMCompiler is omitted from MATH and HumanEval, as its DAG-style planning is not well-suited for problems that require sequential, step-by-step reasoning and tool usage.

\textbf{LLM backend.}
We employed the OpenAI-compatible vLLM (version 0.6.6) server as the LLM serving infrastructure, integrated with PyTorch 2.6 and CUDA 12.8. We enabled \emph{prefix caching}~\cite{vllm}, which reduces redundant computation by reusing previously computed attention states (i.e., \emph{Key-Value cache (KV cache)}) for shared input prefixes across LLM requests. Unless explicitly stated otherwise, all experimental results are obtained with prefix caching enabled. We use Llama-3.1-8B-Instruct~\cite{llama_3.1_8b_instruct} as the default backend LLM. However, to discuss the impact of model size on cost and accuracy, we also use Llama-3.1-70B-Instruct~\cite{llama_3.1_70b_instruct} in \sect{sect:char_implication}.

\textbf{Hardware.}
Experiments were conducted on Google Cloud Platform (GCP). For the 8B model, we used the \texttt{a2-highgpu-1g} instance type with 12 vCPUs (6 physical cores), 85GB memory, and a single NVIDIA A100 40GB GPU\footnote{
In this work, we use GPU-based serving systems for our analysis because they are the de facto standard for large-scale LLM serving. Our characterization methodology and findings are architecture-agnostic and directly transferable to other accelerator platforms like Google TPUs. As detailed in the rest of this paper, key insights such as the impact of agentic control-flow serialization, long-context KV cache pressure, and idle-period underutilization are inherent to the workload characteristics of dynamic reasoning, not to any GPU/TPU-specific microarchitecture. Thus, the system-level implications we identify remain equally relevant to other AI inference accelerators, providing a foundation for future cross-architecture analyses.}. For the 70B model, we used the \texttt{a2-highgpu-8g} instance type with 96 vCPUs (48 physical cores), 680GB memory, and 8 NVIDIA A100 40GB GPUs.
\section{Demystifying AI Agents}
\label{sect:char_workflow}

\sect{sect:workflow_call_count} first examines an agent's \emph{single}-request execution, followed by a detailed exploration of the LLM inference and tool-calling characteristics of agents in \sect{sect:char_llm_inference}. Lastly, \sect{sect:char_serving} shifts the focus to the serving environment of agentic systems where \emph{multiple} requests are handled concurrently, identifying system-level bottlenecks and scalability issues that emerge in agent deployment.

\subsection{Overall Workflow of AI Agents}
\label{sect:workflow_call_count}

\begin{figure}[t!] \centering
\includegraphics[width=0.48\textwidth]{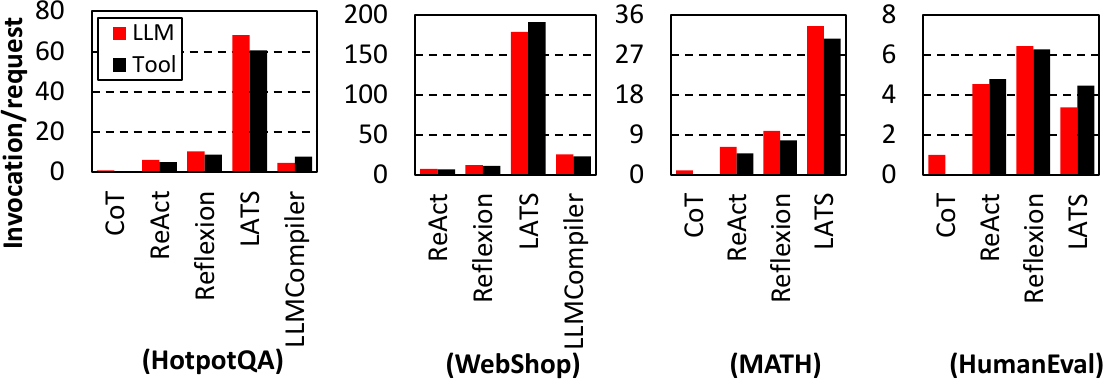}
\vspace{-0.4em}
\caption{
Average number of LLM and tool invocations per request.
}
\label{fig:workflow_call_count}
\end{figure}

{\bf Effect of LLM and tool calls on latency.}
\fig{fig:workflow_call_count} shows the average number of LLM and tool invocations per request across benchmarks. While CoT performs only a single LLM inference per request, tool-augmented agentic systems require significantly more LLM calls, averaging 9.2 times more than CoT. Among these, LATS exhibits the highest LLM invocation count, with an average of 71.0 LLM calls per request. This is primarily due to its use of tree search, which explores multiple reasoning branches (i.e., child nodes) by issuing separate LLM inferences for each one when expanding a tree node.

\begin{figure}[t!] \centering
\includegraphics[width=0.48\textwidth]{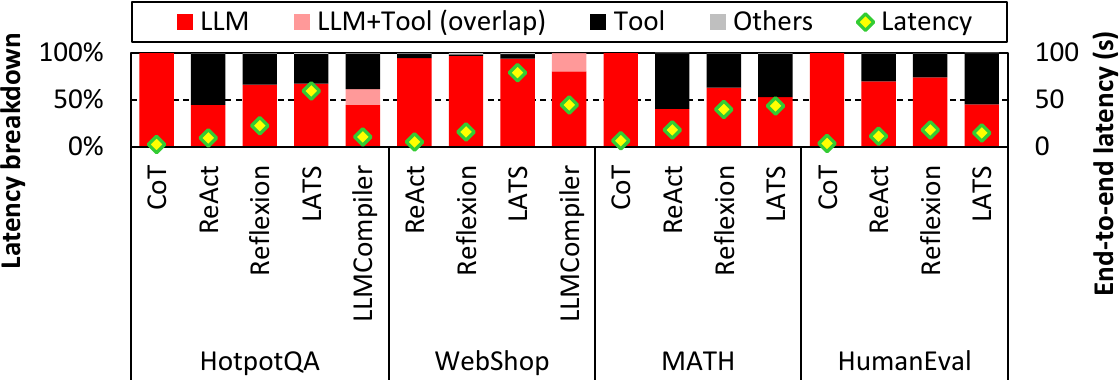}\\
\vspace{-0.4em}
\caption{
Latency breakdown of agents (left axis, bar graph) and their end-to-end latency for processing a single request (right axis, diamond marker). The pink bars represent phases where LLM and tool execution latencies overlap, as observed in LLMCompiler, which asynchronously executes tools during plan generation.}
\label{fig:workflow_latency_breakdown}
\vspace{-0.9em}
\end{figure}

\fig{fig:workflow_latency_breakdown} presents the end-to-end latency and the latency breakdown of each agent's execution. While most agents exhibit a similar number of LLM and tool calls per request (\fig{fig:workflow_call_count}), the latency contribution from tool calls varies significantly depending on the workload. This discrepancy is primarily due to differences in the underlying tool execution latencies. For example, WebShop uses lightweight tools that interact with locally hosted webpages, resulting in tool latencies as low as 20 ms per call. In contrast, HotpotQA relies on the Wikipedia API, where individual calls take an average of 1.2 seconds. As a result, tool execution dominates the overall latency breakdown in this case. 

On average, LLM inference and tool execution account for 69.4\% and 30.2\% of total latency, respectively. Both stages contribute significantly to overall latency, but they are difficult to overlap due to their sequential dependency. Specifically, the LLM output is needed to determine which tool to invoke along with the required arguments. Conversely, the next LLM invocation typically relies on the observation returned by the tool. Although LLMCompiler attempts to mitigate this dependency by streaming intermediate plans to the scheduler for asynchronous execution of tool calls (thus concurrently executing it with planning), the observed overlap accounts for only 18.2\% of total latency.

\begin{figure}[t!] \centering
\includegraphics[width=0.48\textwidth]{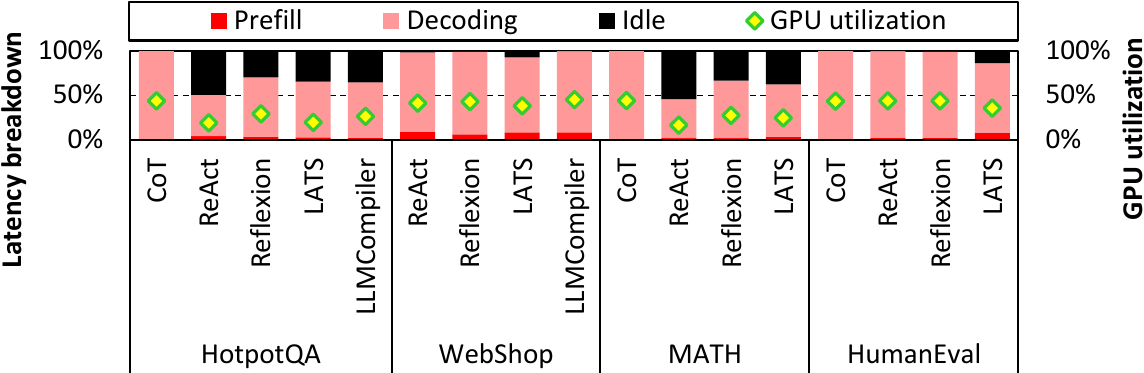}\\
\vspace{-0.4em}
\caption{
Breakdown of GPU runtime by usage (left axis, bar graph) and the resulting average GPU utilization (right axis, diamond marker). GPU utilization is measured as the fraction of actively used GPU cores, using NVIDIA's DCGM~\cite{dcgm}.
}
\label{fig:workflow_gpu_utilization}
\vspace{-1em}
\end{figure}

{\bf Agentic workflow's effect on GPU compute utility.}
\fig{fig:workflow_gpu_utilization} breaks down the GPU runtime by usage and reports the resulting average GPU utilization when handling a single request. Although this setup assumes the processing of a single agent task, concurrent LLM calls can be opportunistically batched to improve GPU utilization, whenever possible, to more efficiently execute agents such as LATS. Unlike CoT, which performs a single LLM inference without external interaction, it is possible for agents to experience longer GPU idle periods due to tool execution. The duration of these idle periods depends on the tool’s latency and whether it leverages the GPU. In WebShop, the tool interacts with locally hosted synthetic web pages, resulting in very short tool latencies (\fig{fig:workflow_latency_breakdown}), so agents do not experience notably higher GPU idle time (i.e., lower GPU utilization). HumanEval exhibits longer tool execution times (\fig{fig:workflow_latency_breakdown}), but the proportion of GPU idle time remains minimal because the tool it calls (which is the test generation tool) utilizes the GPU for LLM execution. In contrast, HotpotQA and MATH employ tools that operate on local CPUs or external systems, leading to substantial GPU idle periods that account for up to 54.5\% of total execution time, resulting in significantly lower GPU utilization compared to CoT.
When the GPU is executing the LLM, its activity can be further divided into the prefill and decode stages, which account for 4.7\% and 74.1\% of the GPU’s execution time, respectively. As noted in ~\cite{sarathi,Leviathan2023SpecDecode,medusa,ainslie2023gqa}, the decode stage is known to be memory-bound. Consequently, the large fraction of time spent in the decode stage further contributes to the underutilization of GPU resources.

Because the sequential dependency between LLM inference and tool calls limits parallel execution opportunities within a single request (i.e., intra-request parallelism), improving overall resource utilization requires leveraging \emph{inter}-request parallelism. We explore this direction in \sect{sect:char_serving}, where we discuss the implications of serving AI agents over multiple queries with LLM request batching~\cite{orca,vllm,vllm_repo}.

\begin{figure}[t!] \centering
\includegraphics[width=0.48\textwidth]{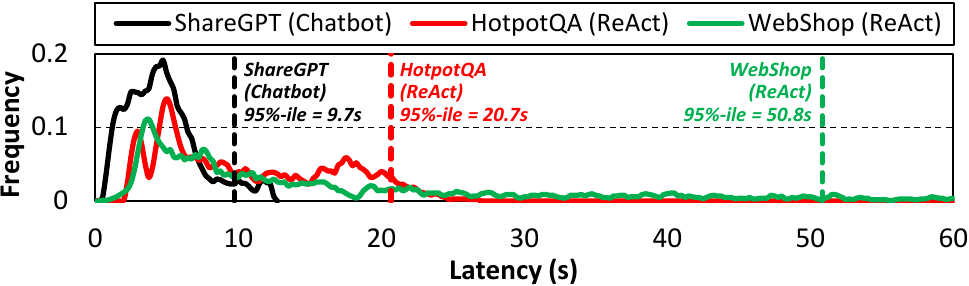}
\vspace{-0.4em}
\caption{
End-to-end latency distribution of a non-agentic ShareGPT workload and ReAct-based agents with HotpotQA and WebShop workloads. Latency is measured while processing one request at a time, with prefix caching enabled.
}
\label{fig:latency_distribution}
\end{figure}

\textbf{End-to-end latency distribution of AI agents.}
\fig{fig:latency_distribution} compares the end-to-end latency distributions of a conventional, non-agentic LLM service using ShareGPT and a ReAct-based agent system using HotpotQA and WebShop.
The ShareGPT dataset represents a typical chatbot workload, where each response is generated by a single LLM inference. 
As shown, this results in a relatively low and consistent latency distribution, with most responses completing within 9.7 seconds. In contrast, the ReAct-based agent exhibits a much broader latency distribution with a heavier tail, due to its multi-step reasoning and reliance on external tools. Because the number of reasoning steps and tool calls varies across requests in agents, the associated computational demands also fluctuate. Consequently, there is significant variance in latency across queries targeting agents.

\subsection{LLM Inference and Tool-Calling Characteristics}
\label{sect:char_llm_inference}

This section further analyzes the behavior of agentic systems by characterizing the properties of LLM inference and tool calls within the AI agent in greater detail.

\begin{figure}[t!] \centering
\includegraphics[width=0.48\textwidth]{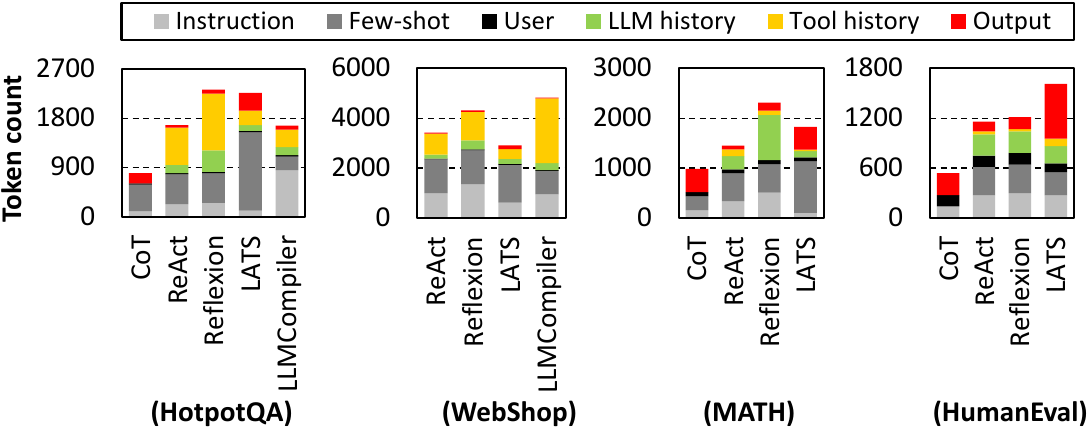}\\
\caption{
Breakdown of input and output tokens in LLM inference. \emph{Instruction} and \emph{Few-shot} (light and dark gray) represent input tokens that are statically fixed as part of the initial prompt to the LLM. \emph{User} (black) denotes input tokens provided by the user as part of the query. \emph{LLM history} (green) and \emph{Tool history} (yellow) represent tokens accumulated from previous LLM outputs and tool responses, respectively, which are then included as input tokens during the next LLM call. \emph{Output} (red) refers to tokens generated by each LLM call.}
\label{fig:token_breakdown}
\vspace{-1em}
\end{figure}

\textbf{Breakdown of input and output tokens in LLM inference.} \fig{fig:token_breakdown} presents the token count distribution across different AI agents. \textit{Instruction} tokens define the agent’s role and objective within the task, while \textit{Few-shot} tokens provide in-context examples that guide the agent’s behavior. \textit{User} tokens represent user queries. \textit{LLM history} and \textit{Tool history} tokens consist of accumulated outputs from previous LLM inferences and tool responses across iterations. \textit{Output} tokens are generated at each LLM inference step, while the remaining tokens collectively make up the input prompt.

Compared to CoT, AI agents generally have longer input tokens. This is because their inputs include additional elements such as agent role-aligned instructions (e.g., LLMCompiler requires instructions to generate a structured plan) and the accumulated context of previous LLM and tool interactions. For output tokens, each LLM call in agent workflows often generates fewer tokens than CoT, except for LATS. This is because agents typically decompose a single task into multiple steps, distributing the overall output across several LLM calls. In contrast, LATS often generates much longer outputs than CoT due to its workflow, where a single LLM call produces multiple candidate samples to expand the tree node.

Token usage patterns also vary depending on the task workload. In knowledge-intensive tasks such as HotpotQA and decision-making tasks like WebShop, tool calls often return large responses (e.g., the full content of a webpage) resulting in longer tool history tokens. In contrast, tasks that rely more heavily on internal reasoning, such as MATH and HumanEval, tend to produce longer LLM-generated outputs, leading to larger LLM history tokens.

Although the ratio of LLM and tool history tokens varies across workloads, most benchmarks exhibit substantial growth in input history over multiple iterations. An exception is LATS, which includes only the path from the root to the current node, rather than concatenating all prior interaction histories. 
In the case of HotpotQA, for instance, initial inputs are typically around 1,000 tokens, but the input size increases to 3–4$\times$ as prior LLM outputs and tool responses are appended to the input context of subsequent LLM calls. Because histories accumulate sequentially, consecutive LLM calls share common prefixes in their input contexts. These long input contexts result in high KV cache usage per request and considerable prefix overlap across iterations. This behavior presents an opportunity to improve GPU compute and memory efficiency through \emph{prefix caching}~\cite{vllm}, as detailed below.

\textbf{Effect of prefix caching on AI agent's compute efficiency.}
Building on the token-level analysis above, we now turn to system-level characteristics, starting with GPU compute efficiency. AI agent workloads involve multiple iterative LLM calls, where a large portion of the input context is reused at each step. Prefix caching leverages this shared prefix to skip redundant computation during the prefill phase by reusing previously cached key-value (KV) pairs.

\begin{figure}[t!] \centering
\includegraphics[width=0.48\textwidth]{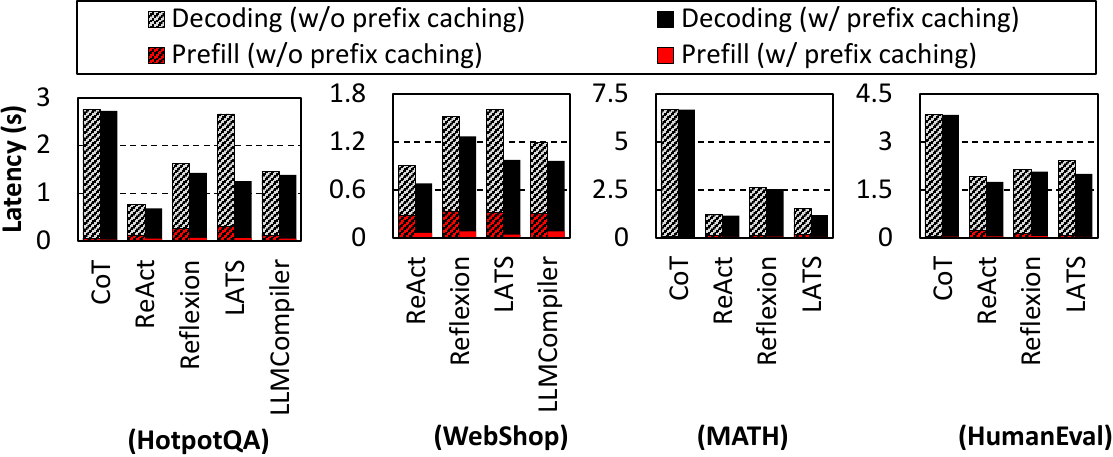}\\
\vspace{-0.4em}
\caption{
Inference latency with(out) prefix caching and its breakdown.}
\label{fig:workflow_llm_latency_and_breakdown}
\vspace{-1em}
\end{figure}

\fig{fig:workflow_llm_latency_and_breakdown} shows LLM inference latency and its proportion of prefill and decoding latency, with and without prefix caching. For CoT, LLM inference occurs only once per request, and the shared prefix across inferences is minimal. Moreover, CoT typically generates a relatively large number of output tokens, making decoding the dominant contributor to latency. In contrast, AI agents operate iteratively and accumulate long input contexts due to interaction histories. As a result, prefix caching reduces prefill latency by an average of 60.1\%, demonstrating its effectiveness in improving compute efficiency by avoiding redundant computations through prefix reuse.
Beyond the prefill phase, prefix caching can also indirectly improve decoding efficiency. In systems that execute multiple parallel LLM calls, decoding may be stalled by ongoing prefill operations. By reducing prefill latency, prefix caching shortens this blocking period, thereby enabling faster decoding and explaining the larger speedup observed in LATS.

Overall, the impact of prefix caching on end-to-end LLM inference latency varies by workload type. While CoT workloads benefit less due to their decoding-dominant property, agentic workloads experience an average 15.7\% reduction in end-to-end latency due to the accumulation of long input contexts over iterative steps. While this per-request improvement may seem modest, the reduction in prefill time can significantly alleviate system-level bottlenecks. In token-level schedulers like vLLM, long prefill phases can delay the scheduling of concurrent requests. By shortening these phases, prefix caching can improve scheduling efficiency and increase overall system throughput. This effect is examined further in \sect{sect:char_serving} (\fig{fig:serving_prefix_caching}).

\textbf{Effect of prefix caching on AI agent's memory efficiency.}
We now discuss the effect of prefix caching on GPU memory requirements by measuring the average GPU memory required to store the KV cache. On average, tool-augmented AI agents consume 3.0$\times$ more memory per request than CoT, and up to 5.4$\times$ more in the worst case. This overhead arises from the iterative nature of agent workflows, where each LLM call appends intermediate reasoning steps and tool responses to the context, resulting in a longer input for each LLM inference.

These results highlight the need for memory optimization in AI agent workloads, with prefix caching serving as a key technique for reducing GPU memory usage. In LATS, multiple LLM inferences are issued in parallel to evaluate several child nodes simultaneously during tree expansion. Without prefix caching, each of these parallel calls creates its own KV cache, resulting in significant memory overhead due to redundancy. With prefix caching, the shared prefix across these parallel calls can be reused, reducing memory requirements by an average of 64.8\% in LATS. For other agents, where all LLM calls are invoked sequentially, prefix caching does not reduce memory usage \emph{within a single request}, since the KV cache cannot be shared across LLM calls. However, in serving scenarios with concurrent requests, prefix caching can significantly improve memory efficiency by reusing the KV cache across requests. We further explore this serving-level memory efficiency in \sect{sect:char_serving} (\fig{fig:serving_memory_efficiency}).

\begin{figure}[t!] \centering
\includegraphics[width=0.47\textwidth]{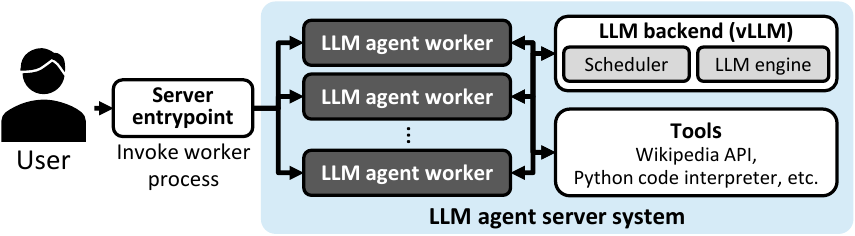}\\
\vspace{-0.4em}
\caption{
High-level overview of our AI agent serving system. 
}
\label{fig:serving_overview}
\vspace{-0.6em}
\end{figure}

\subsection{AI Agent Serving Characteristics}
\label{sect:char_serving}

So far, our characterization has focused on the behavior of AI agents when servicing a single query for a specific task. In this section, we shift our attention to system-level properties of AI agent serving environments, analyzing scenarios where multiple requests are routed to the server and can be processed concurrently for high serving throughput. Unlike static reasoning models that process a user request with a single LLM inference step, AI agents perform multiple reasoning steps iteratively, introducing new challenges for efficient serving.

To examine the characteristics of AI agent serving, we implement an agent serving system, as illustrated in \fig{fig:serving_overview}. When a user sends a request to the agent server’s entry point, each worker processes the request according to the agent’s workflow. Depending on the current step of the task, a worker either sends a request to the LLM inference server or executes a tool. Tool execution may occur locally (e.g., code interpreters, custom functions) or involve external resources (e.g., web search, API calls). Each worker operates asynchronously, and LLM inference requests from multiple workers can be batched at the LLM backend (e.g., vLLM) for high-throughput processing using continuous batching~\cite{orca,vllm}. We adopt vLLM’s default first-come-first-served (FCFS) scheduler in the LLM inference backend. To simulate realistic traffic, input queries to the agent server are randomly sampled and issued to the server following a Poisson arrival distribution~\cite{mlperf_datacenter}.

\begin{figure}[t!] \centering
\includegraphics[width=0.48\textwidth]{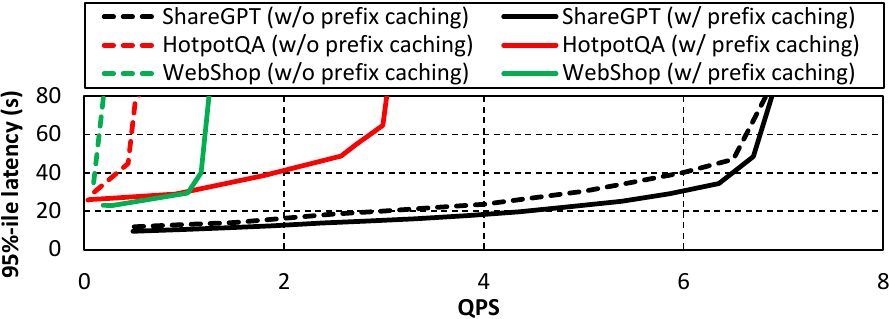}\\
\vspace{-0.4em}
\caption{ 
95th percentile latency for chatbot (ShareGPT) and ReAct-based AI agents (HotpotQA and WebShop) as QPS rates increase, with (solid line) and without (dashed line) prefix caching enabled.
}
\label{fig:serving_prefix_caching}
\end{figure}

\textbf{Importance of concurrent request scheduling.} 
Before comparing the AI agent serving against the conventional chatbot (ShareGPT) serving, we first highlight the importance of concurrently servicing AI agent requests. When ReAct agents are executed \emph{sequentially}, the average latency is 9.6 seconds for HotpotQA and 5.3 seconds for WebShop, limiting throughput to 0.10 and 0.19 queries per second (QPS), respectively. With \emph{concurrent execution}, throughput improves to 2.6 and 1.2 QPS for HotpotQA and WebShop (\fig{fig:serving_prefix_caching}), respectively, achieving 25$\times$ and 6.2$\times$ gains at the cost of a 2.1$\times$ increase in average latency. The greater throughput gain in HotpotQA comes from its longer tool latency, which causes the GPU to remain idle for extended periods. These idle intervals can be effectively utilized by servicing other requests, enabling higher concurrency and throughput.

\textbf{Comparison with conventional static reasoning LLM services.}
We now compare an AI agent serving with a conventional LLM serving scenario, represented by the chatbot (ShareGPT) workload. ShareGPT, a typical single-turn LLM service, processes user queries in a single inference pass. \fig{fig:serving_prefix_caching} shows the changes in end-to-end tail latencies for chatbot (ShareGPT) and ReAct-based AI agent (HotpotQA and WebShop) workloads as input QPS to the server increases. The peak throughput is measured as the maximum sustainable QPS at the knee of the tail latency curve. As depicted, the peak throughput of ReAct is significantly lower than that of ShareGPT. While ShareGPT can sustain up to 6.4 QPS, ReAct supports only 2.6 QPS on HotpotQA and 1.2 QPS on WebShop, even with prefix caching enabled. This limitation stems from ReAct’s multi-step reasoning, where each request involves multiple LLM calls and tool interactions, significantly increasing latency.

\textbf{Effect of prefix caching on AI agent serving throughput.}
Prefix caching is an important system-level optimization that reduces redundant computation during the prefill phase of LLM inference by reusing previously computed key-value (KV) caches. While its impact on the latency of individual LLM calls is modest (\fig{fig:workflow_llm_latency_and_breakdown}), it can substantially improve throughput and serving efficiency for AI agents.

\fig{fig:serving_prefix_caching} compares the effect of prefix caching on chatbot (ShareGPT) and agentic (ReAct) workloads. ShareGPT shows only a modest 1.03$\times$ throughput improvement, as it performs a single LLM call per request with minimal repetition. In contrast, ReAct benefits significantly, achieving an average 5.62$\times$ increase in throughput. This is because agent workloads involve multiple LLM calls per request, amplifying the benefits of avoiding redundant prefill operations.

The performance gap is further explained by token-level batching systems such as vLLM. Without prefix caching, long prefill stages occupy the GPU and block decoding for other requests, leading to system-wide queuing delays. This bottleneck is particularly problematic for AI agents, where repeated LLM calls per request exacerbate inter-request contention. As a result, prefix caching plays a critical role in mitigating these interference effects and improving overall serving efficiency, especially for agentic workloads.

\begin{figure}[t!] \centering
\vspace{-1em}
\captionsetup[subfloat]{captionskip=0.3em}
\subfloat[Average memory usage]{\includegraphics[width=0.22\textwidth]{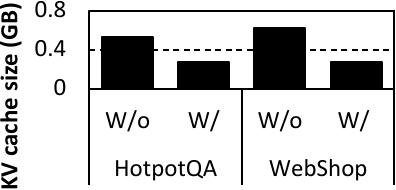}}
\hspace{1em}
\subfloat[Maximum memory usage]{\includegraphics[width=0.22\textwidth]{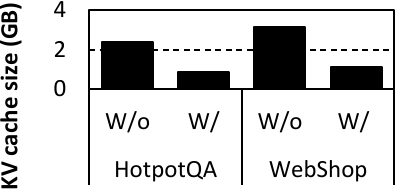}}
\caption{
(a) Average and (b) maximum memory used for KV caches, with and without prefix caching. Evaluation is conducted at 0.2 QPS (HotpotQA) and 0.1 QPS (WebShop) data points using ReAct.
}
\label{fig:serving_memory_efficiency}
\vspace{-0.4em}
\end{figure}

\textbf{Effect of prefix caching on AI agent's memory usage.}
We now investigate the impact of prefix caching on GPU memory efficiency in AI agent serving, focusing specifically on its effect on key-value (KV) cache size, one of the most significant contributors to memory usage in LLM inference. \fig{fig:serving_memory_efficiency} shows the GPU memory consumption for KV cache allocation, with and without prefix caching enabled, under identical QPS conditions. With prefix caching enabled, the average and maximum KV cache memory usage decrease by 51.7\% and 63.5\%, respectively, indicating improved memory efficiency. This reduction arises from the ability of prefix caching to reuse key-value pairs of shared prefix tokens across multiple LLM invocations across AI agent requests. Thus, prefix caching not only improves compute efficiency by eliminating redundant prefill operations but also reduces the KV cache memory footprint, enabling more efficient utilization of GPU memory during AI agent serving.

\section{Demystifying Test-Time Scaling in AI Agents}
\label{sect:char_implication}

We now explore the diverse design space of AI agents and examine their test-time scaling behavior to understand the trade-offs between model accuracy and cost. We evaluated accuracy following the official evaluation protocol of each benchmark. For HotpotQA and MATH, we report exact match accuracy, allowing minor formatting variations (e.g., equivalent mathematical expressions) in MATH. For WebShop, we use the task-specific score defined in the benchmark. For HumanEval, accuracy denotes the proportion of tasks that successfully pass all unit tests. To assess each design point, we used a benchmark of 50 sample questions and measured the average accuracy and the computation cost for each.

\subsection{Analyzing Cost-Efficiency Across AI Agent Design Spaces}
\label{sect:char_design_space}

Deploying AI agents in practical settings requires careful configuration of agentic system parameters. These design choices significantly affect not only the agent’s task success rate but also the overall cost of operating such systems. In this section, we quantify how different parameter configurations in AI agents influence both accuracy and cost-efficiency.

\begin{figure}[t!] \centering
\includegraphics[width=0.48\textwidth]{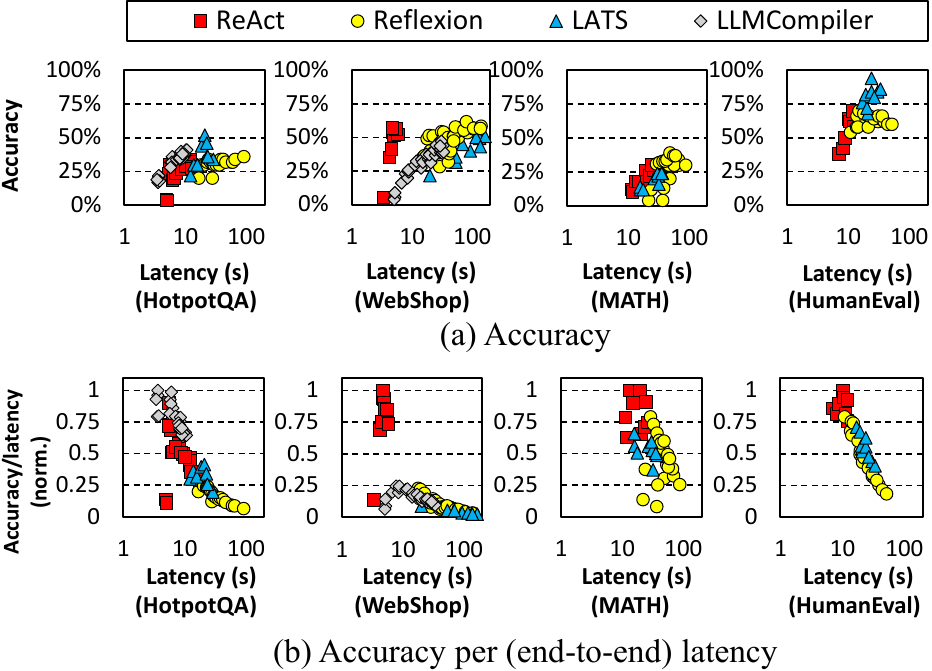}
\vspace{-0.4em}
\caption{
Accuracy and cost-efficiency of AI agent design points. (a) Accuracy vs. latency and (b) Accuracy per latency, illustrating how efficiently each configuration translates cost into task performance.
}
\label{fig:pareto_compute_efficiency}
\end{figure}

\textbf{Pareto analysis of accuracy and cost across AI agent designs.} \fig{fig:pareto_compute_efficiency} presents the trade-off between accuracy and cost across various AI agent configurations. Each point corresponds to a specific design variant, such as changes to the number of few-shot examples or maximum iteration limits.

\fig{fig:pareto_compute_efficiency}(a) shows the trade-off between accuracy and latency. ReAct demonstrates strong compute efficiency across all benchmarks, achieving moderate accuracy with consistently low latency. Reflexion builds on ReAct by introducing reflection steps guided by internal or external rewards. This approach yields modest accuracy improvements but significantly increases latency. LATS extends Reflexion with a tree-based reasoning approach that explores multiple candidate branches at each step. While this leads to higher accuracy, it also introduces substantial computational overhead due to the expansion of reasoning paths. LLMCompiler, with its planning-based architecture, outperforms ReAct on tasks like HotpotQA in both accuracy and cost-efficiency, thanks to its ability to generate and execute structured plans in parallel. However, in tasks such as WebShop—where tool usage involves high interdependencies (e.g., searching or clicking on a webpage)—its DAG-style planning results in unnecessary tool invocations, leading to lower efficiency than ReAct.

\fig{fig:pareto_compute_efficiency}(b) illustrates the cost-efficiency of various agent configurations. We define cost-efficiency as the ratio of accuracy to cost, where cost is measured as end-to-end latency. This metric reflects how effectively each configuration translates compute resources into task accuracy\footnote{Using FLOPs as a proxy for cost (``accuracy per FLOP'') yielded similar qualitative conclusions, so we omit those results for brevity.}. Across all agents and workloads, we observe a consistent pattern: \emph{as computation cost increases, accuracy improves, but with diminishing returns}. This underscores the importance of designing AI agent serving systems that find configurations on (or near) the Pareto frontier, optimally balancing model accuracy against deployment cost rather than optimizing solely for accuracy.

\textbf{Tuning Iteration and Prompting for Cost-Efficient Agent Behavior.}
To better understand the accuracy–cost trade-offs in AI agent design, we analyze how two key parameters in AI agent designs affect model performance: the maximum iteration budget and the number of few-shot examples.

\begin{figure}[t!] \centering
\includegraphics[width=0.48\textwidth]{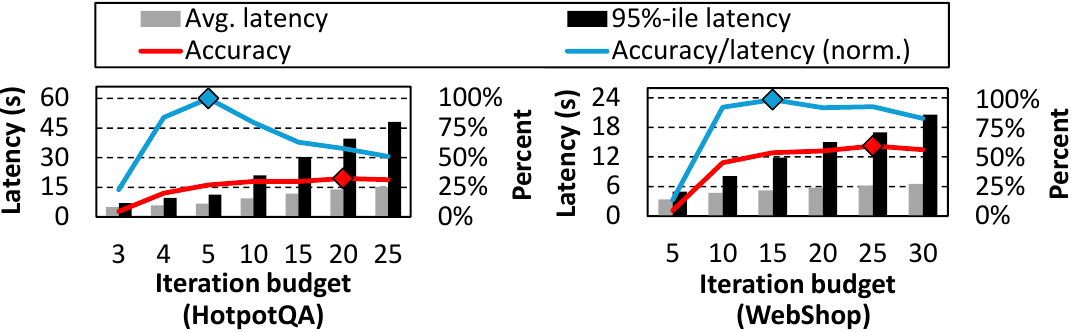}
\vspace{-0.4em}
\caption{
End-to-end latency and accuracy trends under iteration budget constraints in ReAct. 
Markers indicate the points of maximum accuracy (red diamond) and peak cost-efficiency (blue diamond), as measured by accuracy-to-latency ratio.
}
\label{fig:design_space_iteration}
\vspace{-0.8em}
\end{figure}

\fig{fig:design_space_iteration} shows how varying the iteration budget impacts average latency, 95th percentile latency, and accuracy. The iteration budget controls how many reasoning steps and tool invocations the agent is allowed per query. As this budget increases, agents can perform deeper reasoning, which initially improves accuracy. However, both accuracy and average latency eventually saturate, while the 95th percentile latency continues to increase linearly. This rising tail latency is driven by a small set of outlier tasks that consume the full iteration budget. These outliers degrade cost-efficiency by contributing disproportionately to total compute usage without yielding substantial accuracy gains. The widening latency distribution also reduces predictability, which is especially problematic for latency-sensitive deployments. Therefore, iteration limits should be tuned not only for performance but also for latency consistency and operational stability.

\begin{figure}[t!] \centering
\includegraphics[width=0.48\textwidth]{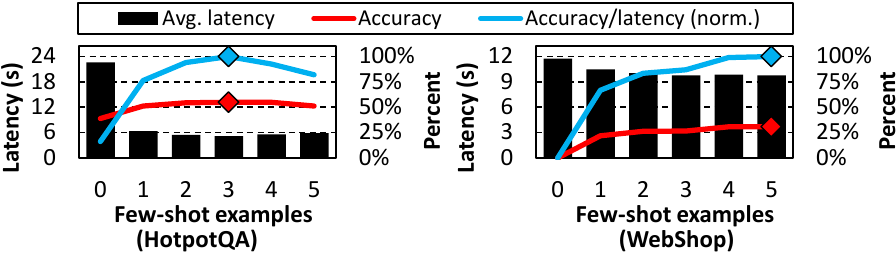}
\vspace{-0.4em}
\caption{
End-to-end latency and accuracy trends with varying numbers of few-shot examples in ReAct. 
Markers indicate the configuration with the highest accuracy (red diamond) and the peak cost-efficiency (blue diamond), based on normalized accuracy-to-latency ratio.
}
\label{fig:design_space_fewshot_latency}
\end{figure}

\fig{fig:design_space_fewshot_latency} shows how varying the number of few-shot examples in the prompt affects latency and accuracy. Initially, adding examples substantially improves accuracy, as agents gain better task understanding. However, beyond a certain point, the benefit diminishes—and in some cases, accuracy declines due to prompt length exceeding the model’s optimal processing range. Interestingly, average latency decreases as more examples are added. This counterintuitive result arises because good examples help agents solve tasks in fewer steps, offsetting the cost of longer prompts. Thus, while longer prompts marginally increase per-token processing time, the reduction in overall reasoning steps often leads to net latency savings. In summary, a small number of carefully chosen examples can improve both accuracy and efficiency, while excessive prompting may lead to diminishing returns.

To identify optimal configurations, we highlight the point at which the accuracy-to-latency ratio is maximized (denoted by blue markers in \fig{fig:design_space_iteration} and \fig{fig:design_space_fewshot_latency}). This point represents the most cost-effective trade-off between model accuracy and response time. Such metrics provide a practical guideline for setting iteration budgets and few-shot prompting under latency or compute constraints.

\subsection{Test-Time Scaling of AI Agents}
\label{sect:char_scaling}

AI agents can dynamically scale their reasoning at test time by adjusting the number of reasoning steps based on task difficulty. This flexibility helps improve performance on complex problems, but it also introduces significant variation in computation cost. Designing systems that are both accurate and efficient requires a deeper understanding of how inference behavior evolves as compute usage increases.

\begin{figure}[t!] \centering
\includegraphics[width=0.48\textwidth]{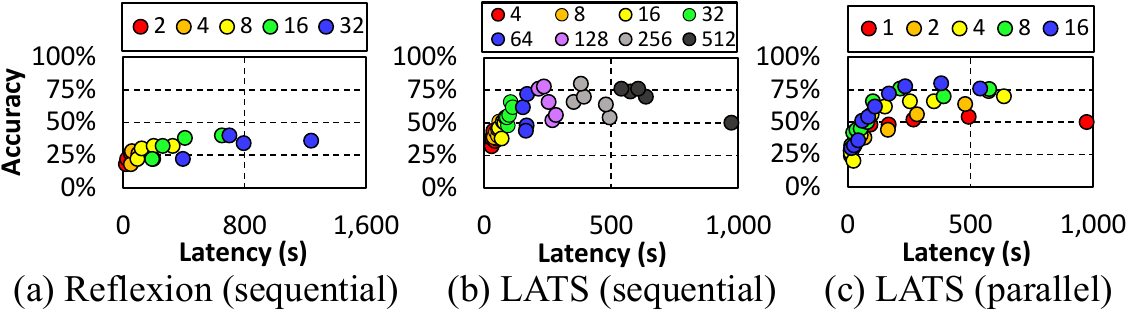}
\vspace{-0.4em}
\caption{
Accuracy-latency trade-offs with sequential, parallel scaling on HotpotQA.
Legends denote the scaling level: maximum reflection steps in (a, b), and number of child nodes per expansion in (c). 
}
\label{fig:test_time_scaling}
\vspace{-0.8em}
\end{figure}

\textbf{Sequential vs. parallel reasoning at test time.}
We investigate the effect of two key forms of test-time scaling for AI agents: \emph{sequential} and \emph{parallel}. In \emph{sequential scaling}, the agent gradually increases its reasoning steps over time, allowing for deeper introspection. This is typical of agents like Reflexion and LATS, where the number of reflection steps can be adjusted dynamically. In contrast, \emph{parallel scaling} issues multiple reasoning branches simultaneously, commonly through parallel LLM calls, to explore diverse solution paths. LATS uses this approach by spawning multiple child nodes during each tree expansion step.

\fig{fig:test_time_scaling}(a) and (b) show the accuracy–latency trade-offs for Reflexion and LATS under sequential scaling. Both methods improve in accuracy with more reflection steps, but with diminishing returns. For example, in Reflexion, increasing latency from 16.9s to 25.6s yields a 4\% accuracy gain. However, achieving the same model accuracy improvement from a later point (56.0s) requires a much larger increase in latency (269.5s), a 31$\times$ higher cost for the same marginal gain.

On the other hand, parallel scaling exhibits a different trade-off. \fig{fig:test_time_scaling}(c) highlights the behavior under parallel scaling in LATS. Increasing the number of child nodes from 1 to 16 improves accuracy by 14.4 percentage points while simultaneously \emph{reducing} latency by 196.3s on average. This is because evaluating multiple reasoning paths in parallel helps the agent converge on high-quality answers more quickly. However, this comes at the cost of issuing more concurrent LLM requests, which increases memory pressure and may limit scalability in multi-tenant or resource-constrained environments.

These results suggest that AI agent configurations should align with system constraints such as latency budgets and available compute resources. Parallel scaling is effective for latency-sensitive workloads, as it allows the agent to explore multiple reasoning paths at once and reach better answers faster. However, it increases resource usage due to the large number of concurrent LLM calls.
In contrast, sequential scaling is better suited for resource-constrained environments. This approach avoids concurrent LLM calls, lowering peak resource demand, but incurs higher latency from step-by-step reasoning.

\begin{figure}[t!] \centering
\includegraphics[width=0.48\textwidth]{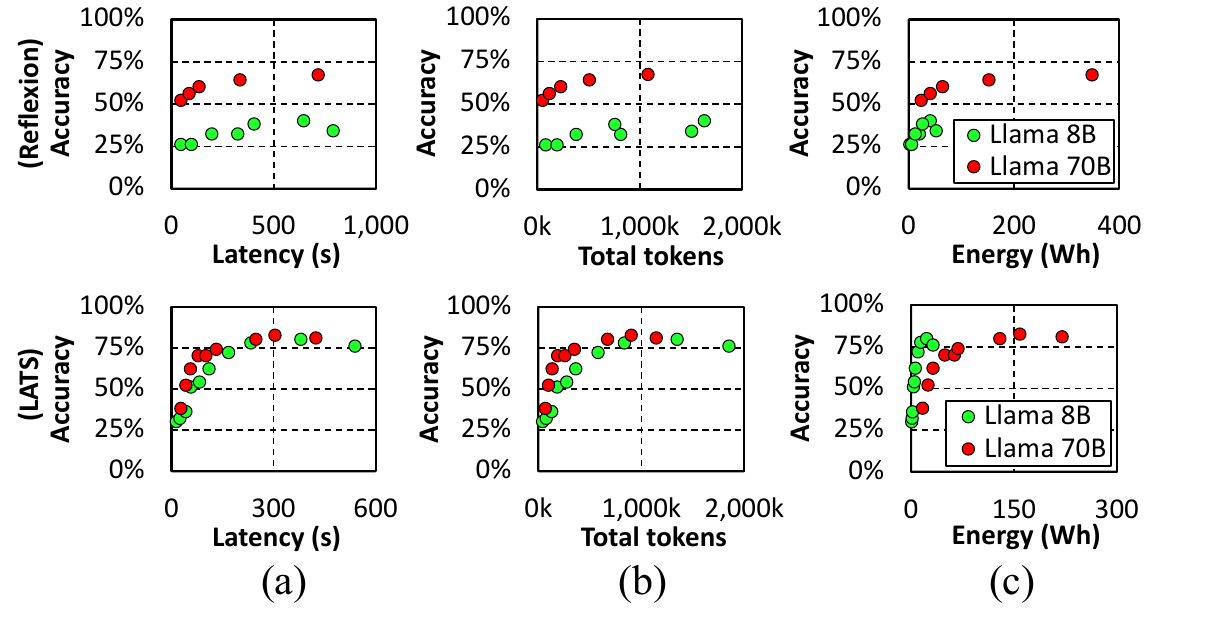}
\vspace{-0.4em}
\caption{
Accuracy–cost trade-offs under test-time scaling across two model sizes (Llama-3.1-Instruct 8B and 70B) on HotpotQA. 
(a)–(c) compare Reflexion (top row) and LATS (bottom row) across latency, token usage, and energy consumption. 
While 70B achieves higher accuracy with fewer steps, the 8B model when paired with parallel scaling can approach 70B performance with lower energy cost. 
Each point denotes a different level of test-time reasoning.}
\label{fig:test_time_scaling_models}
\vspace{-1.0em}
\end{figure}

\textbf{Model size effects on test-time scaling.}
We further analyze how model size affects the accuracy–cost trade-offs under different test-time scaling strategies.

\fig{fig:test_time_scaling_models}(a) shows that both the 8B and 70B Llama-3.1-Instruct~\cite{llama_3.1_8b_instruct,llama_3.1_70b_instruct} models eventually reach saturation in accuracy, but they differ in how quickly they reach this point. The 70B model achieves high accuracy with relatively low latency, whereas the 8B model requires much longer inference times to reach similar performance. This trend is echoed in \fig{fig:test_time_scaling_models}(b), which plots total token usage. The 8B model consumes significantly more tokens at high-accuracy settings, indicating that it needs more reasoning iterations to match the 70B model's performance. However, as shown in \fig{fig:test_time_scaling_models}(c), the 8B model is substantially more energy-efficient. While the 70B model relies on 8 A100 GPUs, the 8B model runs on just one, resulting in lower total energy consumption per request, even when requiring more reasoning steps to be involved.

Interestingly, the performance gap between models can be partially closed with effective scaling strategies. Reflexion (which uses sequential scaling) shows limited accuracy on the 8B model. But with LATS and parallel scaling, the 8B model achieves near-70B performance by exploring multiple paths and selecting the best one. This shows that a test-time strategy can play a compensatory role in low-resource settings.

\section{AI Infrastructure Implications}
\label{sect:ai_infra}

In this section, we analyze the system-level impact of agentic test-time scaling by quantifying the GPU energy consumption and datacenter-wide power demands of AI agents relative to conventional single-turn LLM inference. Following the methodology in \sect{sect:char_scaling}, this section utilizes  Reflexion and LATS as representative AI agents that employ sequential and parallel scaling, respectively. Reflexion and LATS design points were selected based on the highest-accuracy configurations in \fig{fig:test_time_scaling_models}. Llama-3.1-Instruct 8B and 70B models are used as backend LLMs and ShareGPT serves as the baseline for conventional single-turn inference.

\textbf{GPU energy consumption.} 
Reflexion consumes 41.53 Wh and 348.41 Wh per query when using Llama-3.1-Instruct 8B and 70B as backend LLMs, whereas LATS consumes 22.76 Wh and 158.48 Wh (\tab{tab:test_time_scaling_cost}). By contrast, a conventional single-turn LLM inference (ShareGPT) requires only 0.32 Wh (8B) and 2.55 Wh (70B) per query. These figures correspond to a 62.1$\times$–136.5$\times$ increase in GPU energy per query under agent-based test-time scaling (vs. single-turn LLM inference).

Based on recent estimates, ChatGPT serves roughly 500 million to 1.27 billion weekly active users (WAU)~\cite{chatgpt_wau_500M,similarweb_chatgpt_2025,demandsage_chatgpt_stats,forbes_chatgpt_1b_2025}, which corresponds to approximately 71.4 million to 181.4 million daily active users (DAU). Assuming the conservative estimate of 71.4 million DAU and that each user submits just a ``single'' agentic query per day, Reflexion’s daily GPU energy consumption would be approximately 2.97~GWh for the 8B model and 24.89~GWh for the 70B model. Although our analysis does not account for LLM request batching~\cite{orca,vllm}, which can amortize execution overheads, the estimate remains conservative for three reasons: (1) it represents a lower-bound based on the conservative DAU estimate of 71.4 million and assumes just one query per user, despite accelerating adoption and increasing user demand, (2) it includes only GPU energy, omitting overheads from CPU, memory, networking, storage, and cooling, and (3) even the larger 70B model considered in our study is orders of magnitude smaller than today's large-scale LLMs, which now reach hundreds of billions to trillions of parameters~\cite{meta2025llama4,moonshot2025kimi,deepseekai2025deepseekr1,felloai2024gemini15}.

Even under these modest assumptions, the projected demand rivals the daily electricity consumption of Seattle and its surrounding area (24.8 GWh)~\cite{seattle_city_light_report}. As AI agents become increasingly embedded in everyday applications, their query volume could approach, or exceed, that of traditional search engines. For instance, Google Search processes over 13.7 billion queries per day~\cite{google_statistics}, roughly 192$\times$ the 71.4 million agentic queries assumed above. If this growth in user base and usage persists, AI infrastructure demand could rise dramatically, potentially exceeding sustainable limits and underscoring the significant challenges posed by test-time scaling.

\begin{scriptsize}
\begin{table}[t]
\centering
\caption{
Accuracy, latency, and  GPU energy consumption when servicing a single agent request on HotpotQA. Numbers in parentheses indicate the relative increase over ShareGPT (the conventional single-turn inference). 
}
\begin{tabular}{cc|cccccc}
\hline
& & \multicolumn{2}{c}{\makecell[c]{\textbf{Accuracy} \\ \textbf{(\%)}}} & \multicolumn{2}{c}{\makecell[c]{\textbf{Latency} \\ \textbf{(seconds)}}} & \multicolumn{2}{c}{\makecell[c]{\textbf{Energy} \\ \textbf{(Wh/query)}}}\\
\hline
\hline
\multirow{3}{*}{8B} 
    & \textbf{ShareGPT}  & \multicolumn{2}{c}{--} & \multicolumn{2}{c}{4.23 (1×)} & \multicolumn{2}{c}{0.32 (1×)}\\
    & \textbf{Reflexion} & \multicolumn{2}{c}{38} & \multicolumn{2}{c}{649.34 (153.7×)} & \multicolumn{2}{c}{41.53 (130.9×)}\\
    & \textbf{LATS}      & \multicolumn{2}{c}{80} & \multicolumn{2}{c}{380.90 (90.1×)} & \multicolumn{2}{c}{22.76 (71.7×)}\\ \hline
\multirow{3}{*}{70B} 
    & \textbf{ShareGPT}  & \multicolumn{2}{c}{--} & \multicolumn{2}{c}{6.40 (1×)} & \multicolumn{2}{c}{2.55 (1×)}\\
    & \textbf{Reflexion} & \multicolumn{2}{c}{67} & \multicolumn{2}{c}{720.00 (112.6×)} & \multicolumn{2}{c}{348.41 (136.5×)}\\
    & \textbf{LATS}      & \multicolumn{2}{c}{82} & \multicolumn{2}{c}{305.67 (47.8×)} & \multicolumn{2}{c}{158.48 (62.1×)}\\
\hline
\end{tabular}
\label{tab:test_time_scaling_cost}
\end{table}
\end{scriptsize}

\begin{scriptsize}
\begin{table}[t]
\centering
\caption{
Datacenter-wide power demand under current and future traffic scenarios (71.4 Million and 13.7 Billion Queries/day), assuming the HotpotQA benchmark.
}
\begin{tabular}{cc|cccccc}
\hline
& & \multicolumn{3}{c}{\makecell[c]{\textbf{Power @ 71.4 Million} \\ \textbf{Queries/day (Watts})}} & \multicolumn{3}{c}{\makecell[c]{\textbf{Power @ 13.7 Billion} \\ \textbf{Queries/day (Watts})}}\\
\hline
\hline
\multirow{3}{*}{8B} 
    & \textbf{ShareGPT} & \multicolumn{3}{c}{1.0 M} & \multicolumn{3}{c}{182.7 M}\\
    & \textbf{Reflexion} & \multicolumn{3}{c}{123.6 M} & \multicolumn{3}{c}{23.7 G}\\
    & \textbf{LATS}      & \multicolumn{3}{c}{67.7 M} & \multicolumn{3}{c}{13.0 G}\\ \hline
\multirow{3}{*}{70B} 
    & \textbf{ShareGPT}  & \multicolumn{3}{c}{7.6 M} & \multicolumn{3}{c}{1.5 G}\\
    & \textbf{Reflexion} & \multicolumn{3}{c}{1.0 G} & \multicolumn{3}{c}{198.9 G}\\
    & \textbf{LATS}      & \multicolumn{3}{c}{471.5 M} & \multicolumn{3}{c}{90.5 G}\\
\hline
\end{tabular}
\label{tab:power_demand}
% \vspace{-1.6em}
\end{table}
\end{scriptsize}

\textbf{Datacenter-wide power demands.} We now move on to estimating the datacenter-wide power requirements to sustain the aforementioned AI service demands, assuming today's (ChatGPT's 71.4 million queries per day, assuming the conservative ChatGPT DAU estimate and one agentic query per user) and tomorrow's (Google search's 13.7 billion queries per day) AI traffic. \tab{tab:power_demand} translates the per-query GPU energy consumption numbers into datacenter-level power requirements, computed by $P=(\text{Wh/query})\times ((\text{Queries/Day})/(24\text{ hours}))$. Under today’s 71.4 million DAU load, single-turn ShareGPT (70B) requires roughly 7.6 MW, well within the tens-of-megawatts envelope typical of modern datacenters~\cite{hyperscale_datacenter,xAI_collosus}. However, assuming similar traffic levels for AI agents, even the lighter 8B-based agents demand 67.7–123.6 MW, comparable to the power draw of a mid-sized U.S.\ city, while 70B-based agents approach 1 GW, nearly three orders of magnitude higher than the single-turn LLM baseline. Strikingly, this gigawatt-scale power requirement aligns with the announced budget for OpenAI’s multi-gigawatt Stargate facility~\cite{openai_stargate}, which is intended to support \emph{future} AI model deployments. Yet, our analysis suggests that such infrastructure may already be necessary to support agentic systems under today’s traffic levels. Overall, our estimates indicate that even modest user traffic (on the order of tens of millions of queries per day) becomes gigawatt-scale once per-query energy exceeds \(\sim\)100 Wh, a threshold representative of current agentic workloads.

If we were to scale the same per-query figures to Google’s 13.7 billion daily searches, the power numbers would raise single-turn ShareGPT (70B) to 1.5 GW and Reflexion (70B) to nearly 200 GW, far beyond any announced datacenter project (e.g., Meta's recently announced 5 GW AI datacenter Hyperion is scheduled for deployment in 2030~\cite{meta_hyperion_prometheus}) and exceeding the power budgets of many national grids. To put this number into perspective, a 200 GW is almost half of the \emph{entire} U.S. grid's average load (which amounts to 4,178$\times10^3$\text{ GWh}/(365$\times$24$\text{ hours})$$=$ 476.9$\text{ GW}$~\cite{us_grid_load}), a scale usually discussed only for nation-wide decarbonization plans, not for a single industry or technology, one that fundamentally reshapes generation, transmission, and sustainability planning.

{\bf Sustainability challenges of agentic test-time scaling. } Collectively, our findings show that AI agent performance does not scale proportionally with the associated compute, energy, and power costs. Once accuracy saturates, additional test-time scaling yields diminishing returns while imposing substantial system-level burdens. This cost inefficiency is not merely theoretical; it poses concrete constraints on real-world deployments. For instance, OpenAI’s Deep Research~\cite{deep_research}, designed for complex multi-step reasoning, can take up to 30 minutes per request~\cite{deep_research_faq}. To keep infrastructure costs manageable, OpenAI limits usage to 25 runs every 30 days for ChatGPT Plus users~\cite{deep_research_faq}. These limits highlight the financial and computational challenges of sustaining AI systems that rely heavily on intensive test-time computation.

Based on these findings, we argue that building scalable and sustainable AI agents requires moving away from unconstrained test-time scaling. Instead, AI agents should be designed with compute-aware agentic workflows that deliver strong performance through efficient inference, rather than single-handedly relying on extended reasoning depth.

\section{Discussion}
\label{sect:discussion}

\setlength{\fboxsep}{1.2pt}
\textbf{Future directions for sustainable AI agent serving.}
While the primary objective of our work is to raise awareness within the community about the broader system-level implications of deploying agents—particularly the immense infrastructural costs associated with them—we also highlight several promising directions that we believe will be critical for sustainably serving agentic systems.
First, conventional \textit{model-level optimizations} such as quantization~\cite{lin2024awq,xiao2023smoothquant,frantar2022gptq}, distillation~\cite{hinton2015distilling,muralidharan2024compact,sreenivas2024llm,guo2025deepseek}, sparse architectures~\cite{fedus2022switch,yuan2025native,liang2024mixture}, and adaptive model routing~\cite{panda2025adaptive,ding2025best} will remain essential for reducing computational and memory demands. For example, constructing multi-agent systems that combine a heterogeneous mix of small and large language models (SLMs and LLMs), and dynamically selecting the appropriate model depending on the agent’s role and task significance (e.g., planning vs. acting), can substantially reduce both operational cost and latency—an approach also advocated by~\cite{belcak2025small}.
Second, for AI agents that are not strictly latency-sensitive, \textit{carbon-aware computing}~\cite{radovanovic2022carbon,li2023clover,gsteiger2024caribou} that migrates parts of the execution to compute instances incurring lower carbon intensity or electricity costs can provide both environmental and economic benefits.
Finally, \textit{adaptive scaling strategies}~\cite{snell2024scaling,qu2025optimizing,yang2025towards} that dynamically adjust compute resources based on task difficulty and importance enable agents to allocate GPU resources more efficiently, avoiding over-provisioning while maintaining quality of service.
Together, these directions highlight exciting opportunities for improving the efficiency, sustainability, and scalability of serving AI agent workloads.

\textbf{Agent serving under SLA constraints.}
The industry has only recently begun exploring agentic systems, so there are currently no well-established or widely accepted SLA standards. Consequently, we did not (and realistically could not) conduct our analysis with respect to a specific SLA target. The goal of this work is to characterize energy–efficiency trade-offs across diverse agent configurations, rather than to optimize for a fixed latency constraint. Considering the current direction of agentic system development, agents are generally allowed to spend additional computation time to achieve higher reasoning quality. Our analysis intentionally examines this behavior to highlight the inefficiency of unrestricted test-time scaling and to motivate more energy-aware SLA design in future agent deployments. A detailed exploration of efficient agent serving under SLA constraints is left as future work.

\section{Related Work}
\label{sect:related_work}

\textbf{AI agent workflows.} Recent advances in LLM-based AI agents have introduced diverse workflows that combine language-based reasoning with external tool use. Single-agent frameworks (e.g., ReAct~\cite{react}, Reflexion~\cite{reflexion}, LATS~\cite{lats}, LLMCompiler~\cite{llmcompiler}) enhance decision-making through iterative reasoning, tool execution, and reflection. Multi-agent systems, such as CAMEL~\cite{camel} and AutoGen~\cite{autogen}, further extend these capabilities by structuring task execution, communication, and coordinated behaviors among multiple agents. 
While these workflows substantially improve their capabilities and behavioral flexibility, their system-level implications remain underexplored. This work provides the first comprehensive analysis of representative AI agents, offering insights into the efficiency and scalability of agentic systems.

\textbf{AI agent interfaces for tool-augmented reasoning.} In parallel with behavioral advancements in AI agents, recent efforts have focused on standardizing AI agent APIs and protocols to facilitate broader integration and deployment. OpenAI’s function-calling interface~\cite{openai_function_calling} defines a structured mechanism for API invocation, enabling agents to interact with tools in a verifiable and consistent manner. Anthropic’s Model-Context-Protocol (MCP)~\cite{mcp} further formalizes how agents manage context and interact with tools. Google’s Agent-to-Agent (A2A) protocol~\cite{agent2agent} complements these efforts by specifying a standard for multi-agent communication.
Although these contributions primarily standardize the interfaces and protocols for agent interaction, our work takes an orthogonal system-level perspective, uncovering the AI infrastructural challenges posed by agentic workloads under test-time scaling.

\textbf{System-level optimization of AI agents.}  
LLMCompiler~\cite{llmcompiler}, Alto~\cite{alto}, and Ayo~\cite{ayo} reduce inference latency by enabling pipelined and parallel execution across reasoning steps. Autellix~\cite{autellix} optimizes latency through queue-aware scheduling, while AI Metropolis~\cite{ai_metropolis} and Murakkab~\cite{murakkab} improve multi-agent coordination and resource isolation. 
While these works focus on optimizing specific components such as scheduling or execution flow, our study provides a broader characterization of infrastructural behaviors and efficiency trade-offs across diverse AI agents at scale.

\textbf{LLM inference optimization techniques.}
The AI community has only recently begun exploring agents, making it both unclear and highly challenging to determine the most effective methodology for applying various LLM-focused optimization techniques to agentic systems. To maintain clarity and generality in our analysis, this paper focuses on fundamental and widely adopted LLM inference optimizations that are readily available in existing AI serving frameworks~\cite{vllm,zheng2024sglang}. A comprehensive exploration of all the latest LLM optimizations in the literature for agentic systems is beyond the scope of this paper, so we provide a summary of recent LLM inference optimizations and discuss their applicability to agents below. 

In terms of KV cache management, \textit{hierarchical caching}~\cite{cachedattention,infinigen,multiturn} and \textit{non-prefix KV cache reuse}~\cite{cacheblend} approaches extend the naive prefix caching, enabling more efficient KV cache reuse. \textit{Token pruning}~\cite{dejavu,keyformer,attentionsink} or \textit{KV cache compression}~\cite{minicache,kvcachegen}, or model architectural improvement like grouped-query attention~\cite{ainslie2023gqa} and multi-head attention~\cite{liu2024deepseek} reduces the memory footprint of the KV cache, which will be especially helpful for agent workloads with long contexts.  Regarding decoding, \textit{Speculative decoding}~\cite{Leviathan2023SpecDecode} predicts multiple candidate tokens and validates in parallel to reduce decoding latency. In agents, speculative decoding can potentially become effective as their reasoning often generates predictable schema patterns (e.g., JSON structures or function arguments), which will increase the acceptance rate of speculative branches and improve overall decoding throughput. \textit{Prefill-decode disaggregation}~\cite{splitwise,distserve,dynamo} allows flexible and efficient resource allocation by decoupling the compute-intensive prefill phase from the memory-bound decode phase. For agents with long-context that incur substantial prefill computation load, disaggregation mitigates interference between prefill and decoding workloads, leading to more stable performance and improved overall efficiency.

\section{Conclusion}
\label{sect:conclusion}

This paper provides the first system-level characterization of AI agents from an AI infrastructure perspective. While these LLM-based agents demonstrate powerful reasoning capabilities, they also introduce substantial energy overheads that are orders of magnitude higher than conventional single-turn LLM inference. Our analysis shows that common agent design patterns
incur heavy latency penalties and infrastructure costs, especially when deployed at scale. Moreover, test-time scaling yields sharply diminishing returns in accuracy, challenging the cost-effectiveness of current agent implementations.

These findings underscore an urgent need to rethink agent architecture and workflow design. Rather than relying on brute-force test-time scaling, future agents should adopt compute-aware reasoning strategies that optimize accuracy per unit cost. This includes smarter scheduling, caching, prompt engineering, and hybrid scaling approaches that adapt to deployment constraints. By exposing the hidden costs of agentic reasoning and offering actionable insights into their infrastructure impact, we hope this work informs future system and algorithm co-design for scalable and sustainable AI agents.

\section*{Acknowledgment}
This work was partly supported by Institute of Information \& Communications Technology Planning \& Evaluation(IITP) grant funded by the Korea government(MSIT) (No.RS-2024-00438851, (SW Starlab) High-performance Privacy-preserving Machine Learning System and System Software), (No.RS-2024-00395134, DPU-Centric Datacenter Architecture for Next-Generation AI Devices), (No.RS-2025-02264029, Implementation and Validation of an AI Semiconductor-Based Data Center Composable Cluster Infrastructure, 30\%), and by Samsung Research Funding Center of Samsung Electronics (SRFC-IT2402-03). Minsoo Rhu is the corresponding author.

%%%%%%%%% -- BIB STYLE AND FILE -- %%%%%%%%
\bibliographystyle{IEEEtranS}
\bibliography{refs}
%%%%%%%%%%%%%%%%%%%%%%%%%%%%%%%%%%%%

\end{document}